\documentclass[10pt,twocolumn,letterpaper]{article}
            
\usepackage[pagenumbers]{iccv}

\definecolor{iccvblue}{rgb}{0.21,0.49,0.74}
\usepackage[pagebackref,breaklinks,colorlinks,allcolors=iccvblue]{hyperref}
\usepackage{svg}
\usepackage{makecell}
\usepackage{graphicx}
\usepackage{array}
   \usepackage{booktabs} 
   \usepackage{multirow} 

\def\paperID{10}
\def\confName{ICCV}
\def\confYear{2025}

\title{Efficient Self-Supervised Neuro-Analytic Visual Servoing for Real-time Quadrotor Control}

\author{
Sebastian Mocanu\textsuperscript{1} \quad
Sebastian-Ion Nae\textsuperscript{1}\thanks{These authors contributed equally to this work.} \quad
Mihai-Eugen Barbu\textsuperscript{1}\footnotemark[1] \quad
Marius Leordeanu\textsuperscript{1, 2, 3} \\
\textsuperscript{1}National University of Science and Technology POLITEHNICA Bucharest, Romania \\
\textsuperscript{2}Institute of Mathematics "Simion Stoilow" of the Romanian Academy, Romania \\
\textsuperscript{3}NORCE Norwegian Research Center, Norway \\
{\footnotesize\ttfamily\ \{sebastian.mocanu, sebastian\_ion.nae.stud.fils, mihai\_eugen.barbu.stud.acs\}@upb.ro, leordeanu@gmail.com} 
}

\begin{document}
\maketitle

\begin{abstract}
This work introduces a self-supervised neuro-analytical, cost efficient, model for visual-based quadrotor control in which a small 1.7M parameters student ConvNet learns automatically from an analytical teacher, an improved image-based visual servoing (IBVS) controller. Our IBVS system solves numerical instabilities by reducing the classical visual servoing equations and enabling efficient stable image feature detection. Through knowledge distillation, the student model achieves $11\times$ faster inference compared to the teacher IBVS pipeline, while demonstrating similar control accuracy at a significantly lower computational and memory cost.
Our vision-only self-supervised neuro-analytic control, enables quadrotor orientation and movement without requiring explicit geometric models or fiducial markers. The proposed methodology leverages simulation-to-reality transfer learning and is validated on a small drone platform in GPS-denied indoor environments. Our key contributions include: (1) an analytical IBVS teacher that solves numerical instabilities inherent in classical approaches, (2) a two-stage segmentation pipeline combining YOLOv11 with a U-Net-based mask splitter for robust anterior-posterior vehicle segmentation to correctly estimate the orientation of the target, and (3) an efficient knowledge distillation dual-path system, which transfers geometric visual servoing capabilities from the analytical IBVS teacher to a compact and small student neural network that outperforms the teacher, while being suitable for real-time onboard deployment.
\end{abstract}

\section{Introduction}
\label{sec:intro}
Quadrotor applications have expanded from remote controlled aircraft to autonomous systems for aerial photography, mapping, and delivery, typically relying on GPS, IMU, and 9-DOF sensor fusion \cite{arreola2018improvement}.  While effective outdoors, these sensors suffer from degraded performance indoors due to signal interference, multipath effects, and calibration uncertainties \cite{tang2024control, rebbapragada2024c2fdrone}. Vision-based sensors are cheaper and address these limitations through direct environmental perception and immediate visual feedback, enabling real-time feature extraction and spatial awareness for autonomous navigation in GPS-denied environments.

Applications requiring spatial coordination between quadrotors and target objects, such as automated warehouse inspection systems \cite{wawrla2019applications} and dynamic cinematography \cite{nageli2017real}, require robust visual servoing methodologies. In warehouse maintenance scenarios, quadrotors must maintain specific distances \cite{kouris2018learning} and viewing angles relative to infrastructure components while compensating for environmental disturbances \cite{wawrla2019applications}. Similarly, cinematographic applications demand continuous pose adjustment to maintain desired framing as subjects move through complex trajectories \cite{galvane2017automated}.

Visual servoing uses a closed-loop feedback to compare desired image features with current measurements, using feature coordinates and geometric properties as direct control inputs without explicit 3D pose reconstruction \cite{hutchinson1996tutorial, guenard2008practical}. However, classical Image-Based Visual Servoing (IBVS) methods often suffer from numerical instabilities due to singularities in the interaction matrix and conditioning issues during large camera motions \cite{corke1996visual}. While object tracking and pose alignment tasks typically use fiducial markers such as ArUco \cite{kalaitzakis2021fiducial, mraz2020using, cruz2024performance} or AprilTags \cite{olson2011apriltag, zhang2017autonomous, banik2025integrating} that provide easily detectable features with predefined geometric properties, these approaches still face computational challenges and marker dependencies that limit deployment in dynamic, unstructured indoor environments.

Our work presents an efficient marker-free approach utilizing a self-supervised analytical IBVS teacher model that addresses numerical instabilities through reduced classical equations and enhanced image feature detection, coupled with a lightweight student neural network that achieves superior real-time performance. The proposed self-supervised teacher method integrates a two-stage neural network approach: first, a YOLO \cite{redmon2016you, riftiarrasyid2024suitability, yang2025comparative} instance segmentation network for object detection, then a second stage that splits the generated mask in half to estimate the orientation (front/back segmentation) for a toy car target, enabling precise drone positioning without fiducial markers. This knowledge distillation framework transfers the analytical stability of the improved IBVS teacher to a more efficient, low-cost student network while achieving better control performance. System validation follows a progressive methodology, beginning with simulation-based training and subsequently incorporating real-world data for fine-tuning.

The experimental setup requires minimal hardware infrastructure, consisting only of a standard laptop computer and wireless communication link to the quadrotor platform, demonstrating practical feasibility for indoor autonomous operations while also targeting future edge deployment.

\section{Visual-Based Servoing}
\label{sec:related_work}
Visual servoing represents a fundamental control paradigm for mobile robots utilizing camera-based feedback. Classical approaches are categorized into Position-Based Visual Servoing (PBVS) and Image-Based Visual Servoing (IBVS). PBVS reconstructs 3D pose from image features for Cartesian space control\cite{qin2023perception, sun2025towards}, while IBVS directly employs image features as feedback signals, circumventing pose estimation uncertainties\cite{corke1996visual, yang2020autonomous, lee2012adaptive}. For real-time robotic applications, IBVS is predominantly adopted due to its computational efficiency and robustness \cite{yang2024high, kumar2024tracking}.

The IBVS control law is formulated as:

\begin{equation}
\mathbf{v} = -\lambda \mathbf{L}_s^+ (\mathbf{s} - \mathbf{s}^*)
\end{equation}

where $\mathbf{v} = [v_x, v_y, v_z, \omega_x, \omega_y, \omega_z]^T$ represents the camera velocity screw (linear and angular velocities), $\lambda$ is the control gain, $\mathbf{s}$ are current image features, and $\mathbf{s}^*$ are desired features.

The interaction matrix $\mathbf{L}_s$ (image Jacobian) relates feature motion to camera motion:

\begin{equation}
\dot{\mathbf{s}} = \mathbf{L}_s \mathbf{v}
\end{equation}

For each point at pixel coordinates $(u, v)$ with depth $Z$, the interaction matrix is:

\begin{equation}
\label{eq:ibvs-full}
\mathbf{L}_s = \begin{pmatrix}
-\frac{f}{Z} & 0 & \frac{\bar{u}}{Z} & \frac{\bar{u} \bar{v}}{f} & -\frac{f^2 + \bar{u}^2}{f} & \bar{v} \\
0 & -\frac{f}{Z} & \frac{\bar{v}}{Z} & \frac{f^2 + \bar{v}^2}{f} & -\frac{\bar{u} \bar{v}}{f} & -\bar{u}
\end{pmatrix}
\end{equation}

where $\bar{u} = u - u_0$ and $\bar{v} = v - v_0$ are the pixel coordinates relative to the principal point, while $f$ is the focal length expressed in units of pixels.

The pseudo-inverse $\mathbf{L}_s^+$ is computed as:

\begin{equation}
\mathbf{L}_s^+ = (\mathbf{L}_s^T \mathbf{L}_s)^{-1} \mathbf{L}_s^T
\end{equation}

Feature detection represents a critical component for IBVS implementation. Marker-based approaches predominate due to their reliability and implementation simplicity. Inria demonstrated ArUco marker tracking using a Parrot Bebop 2 for moving target applications\cite{Marchand05b} and PID-based tracking of moving targets\cite{teuliere2011chasing}. Similar marker-based systems have been deployed for autonomous drone landing and surface detection tasks\cite{keipour2022visual, drones8110605}, establishing their efficacy in real-world scenarios.
However, marker-based methods face practical limitations including sensitivity to environmental conditions such as variable lighting, partial occlusion, and marker degradation. Additionally, marker installation and maintenance requirements limit applicability in dynamic or inaccessible environments \cite{liang2020moving}.

Recent advances in deep learning have introduced alternative feature detection approaches. Neural networks as keypoint detectors have been employed for specific object tracking, as demonstrated by\cite{sepahvand2025deep}, where a CNN was trained to detect keypoints on rectangular objects under varying illumination and occlusion conditions. Other works focus on bounding box detection and object center tracking during aggressive maneuvers\cite{10140151}.
The YOLO family of models has been extensively adopted for robotics applications\cite{li2023design, zhang2024vision}, including instance segmentation\cite{mohamed2021insta, tang2024visual}, object detection\cite{drones8110605}, and image classification\cite{zhao2024real}, demonstrating the versatility of deep learning approaches for visual servoing feature extraction \cite{pathre2024imagine2servo}.

Recent approaches integrate neural networks with visual servoing algorithms to enhance robotic control. Liu et al.\cite{liu2019image} developed a two-stream network for robotic arm manipulation, directly outputting velocity commands to achieve desired poses without explicit visual servoing computations. Knowledge distillation techniques have shown promise in lightweight servo networks, as demonstrated by\cite{drones8110605}, where a student network learns optimal grasp point identification for tree branch manipulation.

Our work extends these concepts by distilling knowledge from YOLO segmentation networks combined with IBVS algorithms, eliminating fiducial marker requirements while maintaining computational efficiency. The proposed lightweight network directly generates velocity commands from single camera frames. Additionally, we address the simulation-to-reality gap through continuous learning, initially training in simulation and subsequently fine-tuning with real-world data to ensure robust performance across deployment environments.

\section{Our Approach}
\label{sec:proposed_method}
\begin{figure*}[t]
    \centering
    \includegraphics[width=\textwidth]{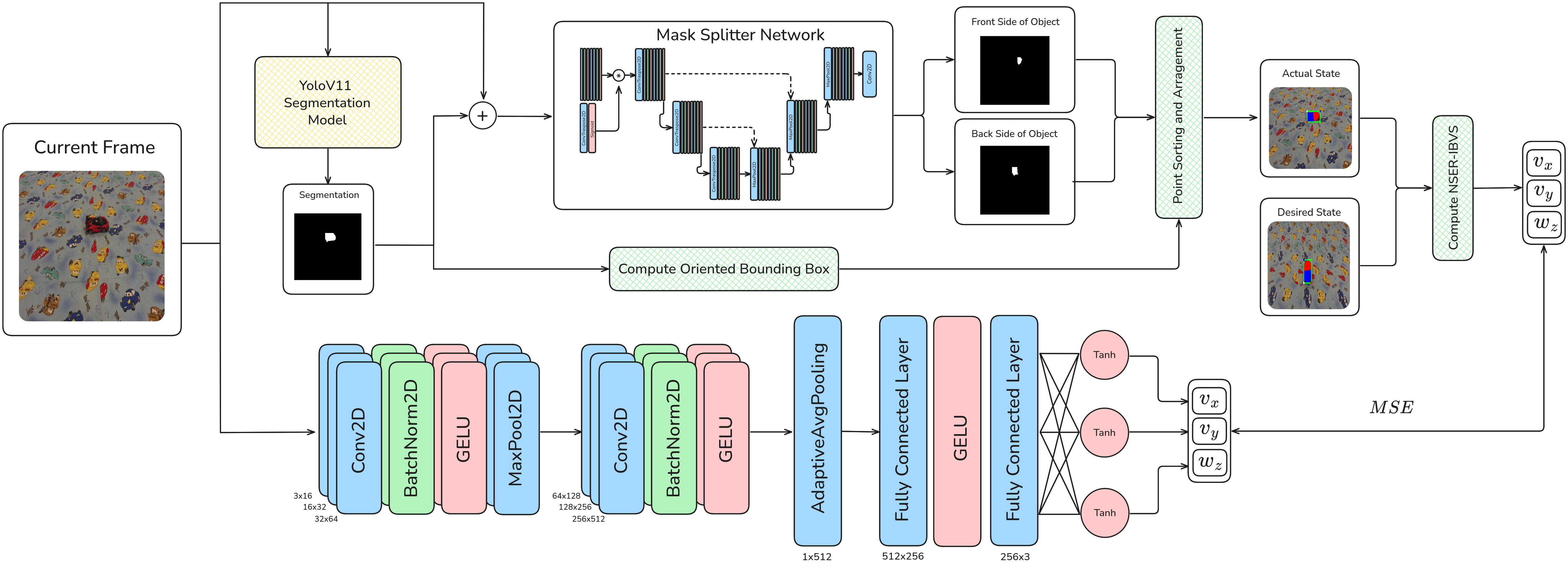}
    \caption{Overview of the proposed Teacher-Student Self-supervised Neuro-Analytic model. Top row illustrates the NSER-IBVS (analytical) Teacher path, which uses YOLO for object segmentation and a mask splitter network to estimate object orientation relative to the drone camera. The bottom one represents the Student path, which learns by minimizing the MSE cost between its output drone commands ($\nu_x$, $\nu_y$, $\omega_z$) and the Teacher's output.}
    \label{fig:teacher-student-math}
\end{figure*}

Our method consists of a teacher-student framework for vision-based quadrotor control, both teacher and student are tested in a digital-twin of the real-world environment. 

The teacher employs a two-stage neural network processing system that is combined with an image-based visual servoing algorithm. In the first stage, a small YOLOv11 Nano (2.84M parameters) instance segmentation network is used. This network is initially trained on synthetic data from the digital-twin and subsequently fine-tuned using real images. This stage has two main objectives: to compute an oriented bounding box (OBB) of the car and to provide its mask to the second stage of our system, as the OBB alone is insufficient for accurate pose inference. The second stage is a specialist neural network trained to split the YOLO masks in half, generating masks that represent the front and back regions of the target vehicle. Knowing the position of these front and back masks and having the OBB, we can rearrange the keypoints to generate four ordered points (two per region) that serve as stable visual features for IBVS. The image Jacobian and control velocities are computed using these features to minimize the error and achieve the desired relative state. The desired orientation keypoints are established by capturing the target pose at the beginning, applying segmentation, and extracting the corresponding feature points. Then we preprocess these  keypoints for stability enhancement, detailed in Section \ref{sec:ibvs_implem}. Our method then rectifies its pose with respect to the desired reference pose.

This framework constitutes the teacher model, which is distilled into a compact convolutional neural network trained via regression to output quadrotor control velocities from camera input RGB images detailed in Section \ref{sec:know_dist}.

\begin{figure*}[t]
    \centering
    \includegraphics[width=\textwidth]{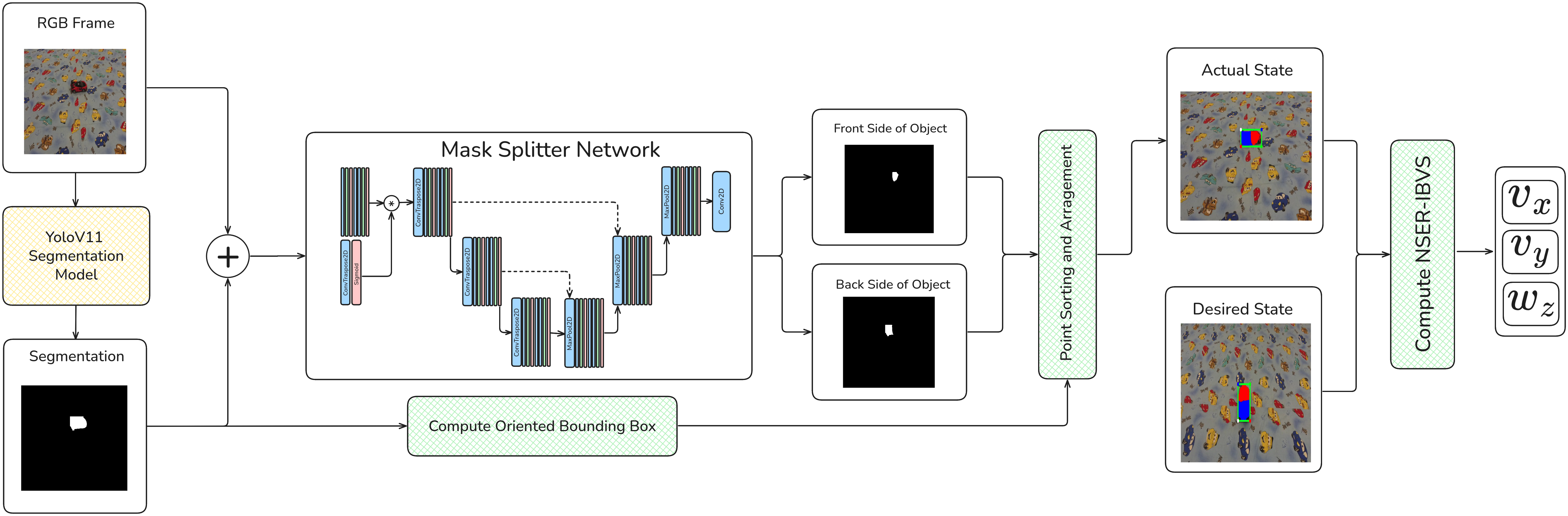}
    \caption{Overview of the proposed Numerically Stable, Efficient, Reduced Image Based Visual Servoing (NSER-IBVS) architecture. The pipeline takes an RGB frame, generates segmentation, and feeds them as input to the Mask Splitter Network. From the original segmentation, bounding box coordinates are determined and used together with the divided front and back masks to rearrange coordinates positions. These coordinates serve as visual features for the analytical model to compute the velocity commands ($\nu_x$, $\nu_y$, $\omega_z$).}
    \label{fig:nser-ibvs}
\end{figure*}

\subsection{Numerically Stable Efficient Reduced IBVS}
\label{sec:ibvs_implem}

\begin{figure}[!ht]
    \centering
    \begin{minipage}[b]{0.31\linewidth}
        \centering
        \includegraphics[width=\linewidth]{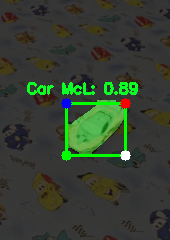}
    \end{minipage}
    \hfill
    \begin{minipage}[b]{0.31\linewidth}
        \centering
        \includegraphics[width=\linewidth]{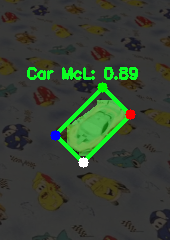}
    \end{minipage}
    \hfill
    \begin{minipage}[b]{0.31\linewidth}
        \centering
        \includegraphics[width=\linewidth]{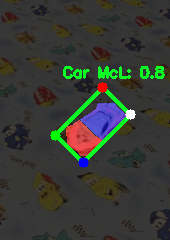}
    \end{minipage}
    \caption{Comparison of different bounding box approaches derived from segmentation results: (left) regular bounding box including parts that do not belong to the object, (middle) oriented bounding box that may vary in orientation between frames, and (right) oriented bounding box using a mask-splitting network to separate anterior and posterior vehicle components for improved ordering stability.}
    \label{fig:bb-vs-obb-yolo-seg}
\end{figure}

The proposed method begins by training a YOLOv11 Nano instance segmentation model using simulation-generated data, forming the foundation of a two-stage segmentation pipeline. The first stage segments the entire vehicle, while a secondary network splits the segmented region into anterior and posterior components, as illustrated in Figure \ref{fig:nser-ibvs}. While effective for segmentation, the first stage generates bounding boxes that may include parts not belonging to the object. To address this issue, we compute a bounding box around the segmented object mask. This produces an unstable point ordering at inference time, introducing inconsistencies for the analytical IBVS algorithm. By determining the positions of the frontal and rear masks, we can rearrange these points to create an oriented bounding box that better represents the car pose and provides more stable feature points, as shown in Figure \ref{fig:bb-vs-obb-yolo-seg}.

The mask splitting network employs a U-Net \cite{ronneberger2015unetconvolutionalnetworksbiomedical} architecture with attention mechanisms \cite{oktay2018attentionunetlearninglook}, accepting a 4-channel input tensor that comprises the original RGB image and the binary vehicle mask from YOLOv11. The encoder-decoder structure features skip connections and progressively downsamples through three stages with channel dimensions of 32, 64, 128, and 256. An attention mechanism applied to the mask channel generates spatial weights that modulate image features to focus on vehicle regions. The decoder reconstructs spatial resolution through transposed convolutions, producing a 2-channel output that represents anterior and posterior segmentation masks. The network contains approximately 1.94M parameters and incorporates dropout regularization in deeper layers preventing overfit.

The splitter loss function partitions a binary segmentation mask into complementary front and back regions by reconstructing the original input mask. Multiple constraints enforce proper partitioning: individual binary cross-entropy losses ensure that each predicted mask matches its ground truth target, a partition constraint penalizes deviations when the sum of both predicted masks differs from the original mask, and overlap penalties discourage simultaneous occupancy of the same pixels. The coverage loss ensures accurate reconstruction by penalizing both excess predictions beyond the original mask and the missed areas within it. Loss weights are dynamically scheduled during training, emphasizing individual mask accuracy, then gradually shifting toward stronger partition and overlap constraints.

The resulting segmentation masks are processed to compute an oriented bounding box by fitting the minimum enclosing rectangle to the object boundary. The four corner coordinates of this bounding box serve as visual features of the IBVS framework. To ensure keypoint stability and consistent ordering, we employ a preprocessing algorithm that assigns points to anterior or posterior regions based on their proximity to the respective mask centroids. The points are subsequently ordered clockwise to maintain temporal consistency across frames, thereby preventing feature correspondence ambiguity that could destabilize the IBVS control loop.

Control velocity computation is achieved by comparing current keypoint coordinates with reference features obtained from a desired pose configuration. Due to the underactuated nature of the quadrotor platform, we use an efficient approach to compute only the necessary velocity components from the classical IBVS controller by eliminating the angular velocity components $\omega_x$ and $\omega_y$ and the linear velocity $\nu_z$. Furthermore, since our controller operates exclusively at fixed altitude, we also remove the linear velocity component $v_z$. 

Starting from Equation \ref{eq:ibvs-full}, the reduced interaction matrix takes the form:
\begin{equation}
\label{eq:numerical-stable-efficient-reduced-ibvs}
\begin{pmatrix}
\dot{u} \\
\dot{v}
\end{pmatrix} = \begin{pmatrix}
-\frac{f}{Z} & 0 & \bar{v} \\
0 & -\frac{f}{Z} & -\bar{u}
\end{pmatrix} \begin{pmatrix}
v_x \\
v_y \\
\omega_z
\end{pmatrix}
\end{equation}

that we use to stack the Jacobians corresponding to the keypoints from the oriented rectangle.

\subsection{Teacher-Student Model Distillation}
\label{sec:know_dist}

To enable real-time deployment, we employ knowledge distillation \cite{hinton2015distillingknowledgeneuralnetwork} to transfer the visual servoing capabilities from the teacher IBVS pipeline to a lightweight, low-cost student network. The student network is designed as a 1.7M-parameter CNN that directly regresses control velocities from RGB images using mean squared error (MSE) loss.

A key component of this architecture is the normalization of the target labels (the velocities commands) and their respective de-normalization at inference time. To train this student network, we require both RGB images and their corresponding control velocities. We analyzed the values from the experiments conducted on both real-world and digital-twin environments and determined the minimum and maximum bounds for each command across all scenes, as shown in Table \ref{tab:cmd-values}. The normalization of target labels improves training stability by ensuring all velocity commands operate within similar numerical ranges, preventing certain outputs from dominating the loss function due to scale differences. This approach also enhances gradient flow during backpropagation and enables the use of bounded activation functions like tanh, which naturally constrains the network outputs to physically meaningful control ranges while maintaining differentiability throughout the training process.

\begin{table}[h!]
    \centering
    \begin{tabular}{lcc|cc|cc}
        & \multicolumn{2}{c}{$\nu_{x}$} & \multicolumn{2}{c}{$\nu_{y}$} & \multicolumn{2}{c}{$\omega_{z}$} \\
        \cmidrule(lr){2-3} \cmidrule(lr){4-5} \cmidrule(lr){6-7}
        & min & max & min & max & min & max \\
        \hline
        Sim  & -24 & 16 & -9 & 9 & -27 & 40 \\
        Real & -30 & 30 & -30 & 30 & -40 & 40 \\
        \bottomrule
    \end{tabular}
    \caption{Minimum and maximum command values for translational velocities $\nu_{x}$ and $\nu_{y}$ and rotational velocity $\omega_{z}$ in digital-twin (Sim) and real-world (Real) collected from NSER-IBVS evaluation runs. Used to determine the normalization of the target labels for the self-supervised method.}
    \label{tab:cmd-values}
\end{table}

The student architecture consists of a convolutional feature extractor followed by fully connected regression layers. The convolutional backbone progressively increases channel dimensions from 16 to 512 through six convolutional blocks, each incorporating batch normalization and GELU activation functions. Max pooling is applied after the first two blocks for spatial downsampling, while the final layers use kernel sizes of 3x3 to capture fine-grained spatial features. An adaptive global average pooling layer reduces spatial dimensions to 1x1, followed by two fully connected layers (512→256→3) that output final velocity commands.

Given the limited availability of real-world data, training is performed in two stages. The first stage is pretraining on simulated trajectories from the digital-twin. The second stage is fine-tuning on real-world data. For the real-world fine-tuning stage, we implement targeted data augmentation strategies to enhance dataset diversity and improve generalization to unseen conditions.

The distillation process transfers knowledge from the NSER IBVS teacher to the neural student, enabling direct end-to-end control from visual input while maintaining the characteristics of the classical visual servoing approach, thereby yielding similar trajectories and behavior.

\subsection{Data Generation and Simulation Environment}
\label{sec:dataset_sim}

The experimental environment is replicated using a simulator based on the Parrot Anafi drone platform with its accompanying Sphinx simulation framework \cite{WhatisPa31:online}. This simulation represents a digital-twin for our real-world testing environment, featuring accurate measurements, object structure and poses. The physical test setup consists of a $5\times4$-meter carpet with a centrally positioned toy car target and the quadrotor positioned at varying poses from the object. The carpet is necessary because the bunker-like environment has a non-Lambertian floor, which poses challenges for our UAV to maintain a fixed position. 

Simulated environment creation uses Unreal Engine 4 for level design and asset placement and Blender \cite{blender2024} to create the meshes for most of the assets. Since the target vehicle was harder to reproduce in Blender, it was digitized using Hunyuan3D-2 \cite{zhao2025hunyuan3d20scalingdiffusion} to generate a \texttt{.fbx} mesh file, which is then scaled to match the dimensions of the physical objects in the real-world for accurate simulation fidelity.

Data generation involves flight sequences within both the digital-twin and real-world environments using varying gimbal camera angles ($30^\circ$, $45^\circ$, $60^\circ$, $75^\circ$, $90^\circ$) for training and $45^\circ$ for validation, since that is the angle we want to test our method. For the simulated environment, we used two rendering configurations: low and high quality that affect lighting and mesh fidelity to ensure generality and a wide-range of FOVs. With this approach, we created a dataset of $14693$ frames for training and $1123$ for validation from the digital-twin environment, and $13760$ frames for training and $1084$ frames for validation from the real-world dataset. The frames were subsequently augmented to enhance generalization capability.

The collected data serves as input for both the YOLO segmentation and Mask Splitter pipelines employed for object segmentation and keypoint orientation reassignment. For the YOLO model, we annotated the data using polygons over $9302$ frames for simulated environment and $8637$ for the real one. After training the segmentation model, for the Mask Splitter we implemented a simple labeling tool which divides the car segmentation based on the user interaction. The algorithm first calculates the centroid of the car mask using image moments and then prompts the user to click around the front portion of the vehicle. Using the centroid as reference point, it creates a direction vector from the center to the user-selected front point and projects all mask pixels onto this vector using dot product calculations. Pixels with positive dot products lie in the same direction as the front point relative to the centroid and are assigned to the front mask, while the negative dot products are assigned to the back mask, effectively creating a linear split that separates the vehicle into front and rear segments.

\section{Experiments}
\label{sec:experiments}
\subsection{Experimental Setup}
\label{subsed:experiments-setup}
\begin{figure}[!ht]
    \centering
    \includegraphics[width=0.8\linewidth]{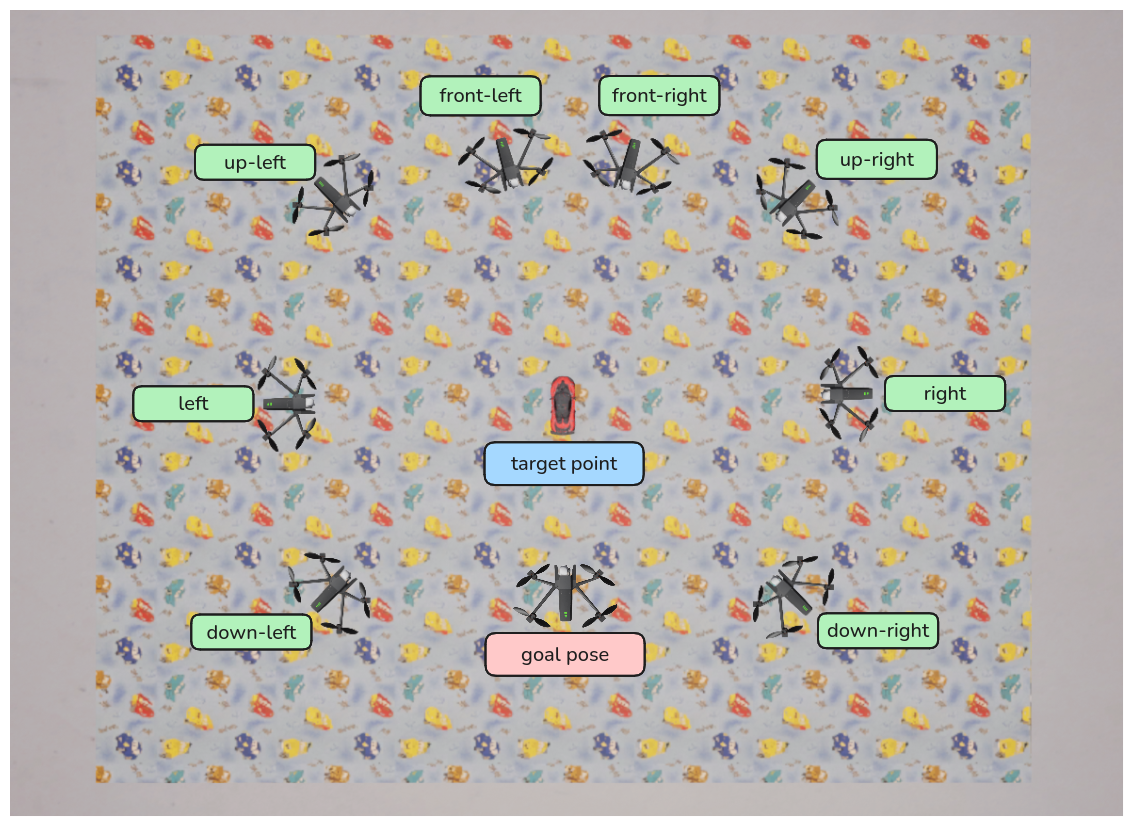}
    \caption{Environment configuration showing 8 starting poses around the target car for the flight tests. The drone must navigate from each start pose to the goal pose exactly behind the car, while keeping the car within the camera frame. The up-left, up-right, down-left and down-right poses are oriented at $\pm45^{\circ}$ relative to the car, while front-left and front-right poses are at $\pm14^{\circ}$.}
    \label{fig:drone-starting-poses}
\end{figure}

To verify our system's performance, we systematically sampled 100 simulation runs for each viewpoint across eight distinct drone poses, yielding 800 total evaluations (see Figure \ref{fig:drone-starting-poses}). We applied both teacher and student methods to these drone poses, generating the same number of evaluations for each approach.

The student model was trained on just 30 simulation runs across each of the presented drone poses, totaling 240 training run samples ($74307$ frames). We applied data augmentation using a multiplication factor of 5 (4 augmented + 1 original), creating synthetic diversity through randomized saturation, brightness shifts, and salt-and-pepper noise at each epoch while preserving one set of original frames. The velocity commands in the dataset were normalized across $v_x$, $v_y$, and $\omega_z$ according to the data distribution (Table \ref{tab:cmd-values}), and images were resized to $224 \times 224$ pixels. Since our objective is to output velocity commands, this resizing does not affect the method's performance. Training was performed using Adam optimizer with a learning rate of 0.001 and mean squared error as the loss function. We monitored validation loss using an early stopping mechanism with a patience of 3 and $\ \delta=10^{-4}$.

\begin{figure*}[!ht]
    \centering
    \begin{minipage}[t]{0.48\textwidth}
        \centering
        \includegraphics[width=\linewidth]{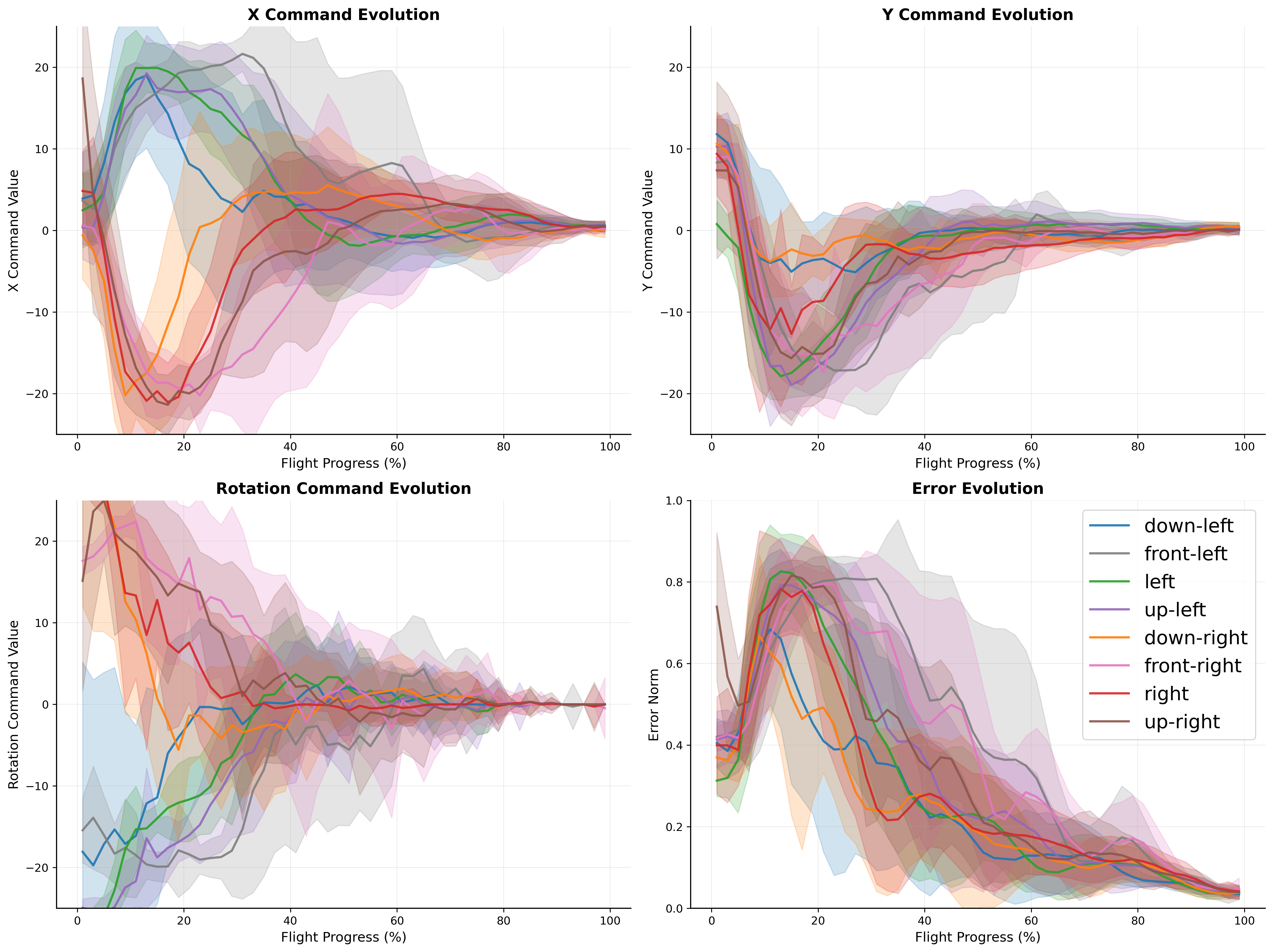}
        \caption*{(a) Real-world analytical teacher (NSER IBVS): evolution of drone control commands and tracking errors across flight trajectory.}
        \label{fig:real-ibvs-evolution-over-time}
    \end{minipage}
    \hfill
    \begin{minipage}[t]{0.48\textwidth}
        \centering
        \includegraphics[width=\linewidth]{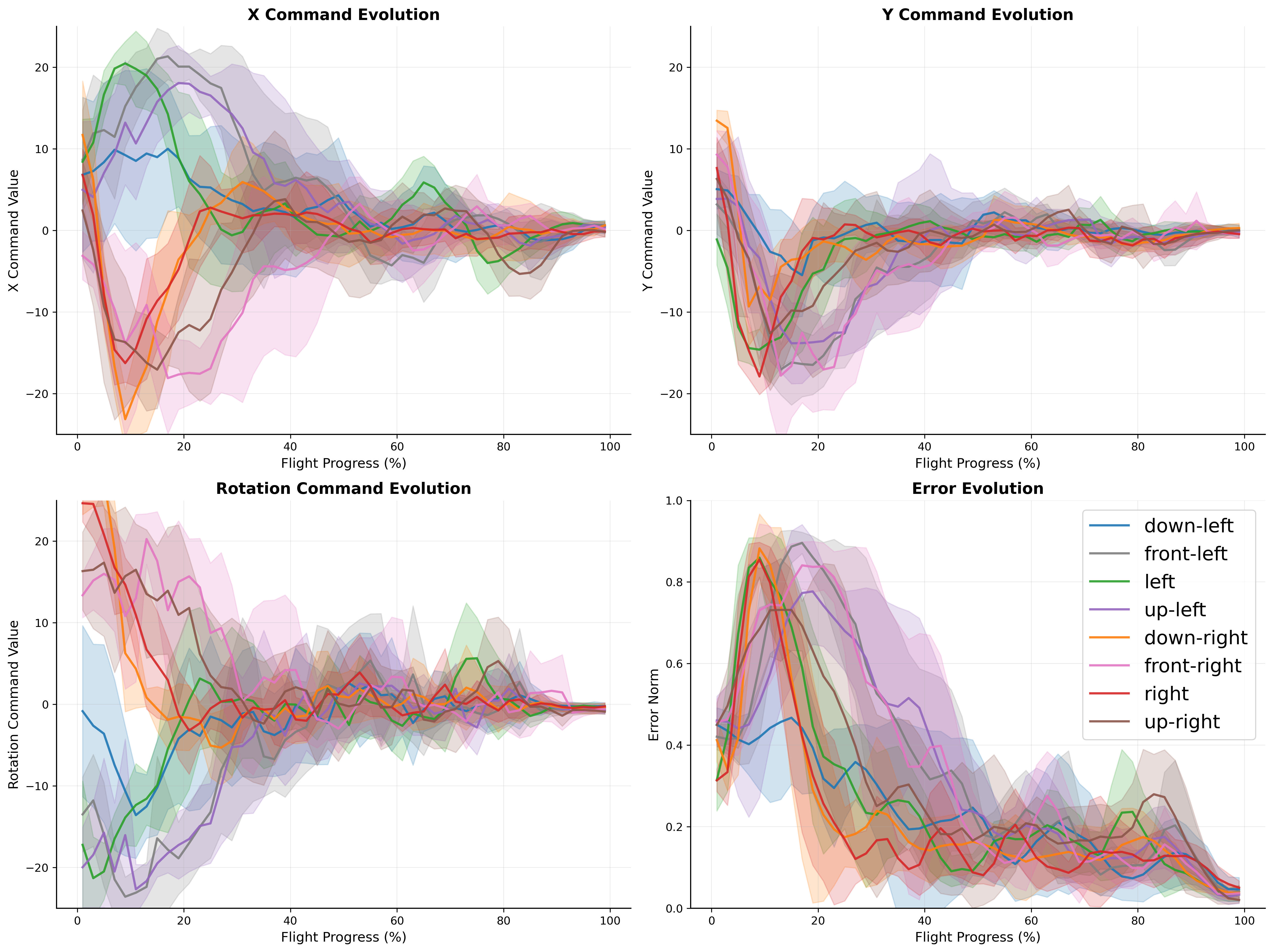}
        \caption*{(b) Real-world small ConvNet student, which distills the analytical teacher: evolution of drone control commands and tracking errors across flight trajectory.}
        \label{fig:real-student-evolution-over-time}
    \end{minipage}
    \caption{Comparison of drone control command and error evolution on novel test sequences for (a) teacher (NSER IBVS) and (b) student (Self-Supervised Neuro-Analytical) across flight trajectories from 8 different starting points (see Fig.~\ref{fig:drone-starting-poses}). Solid lines represent mean values; shaded areas indicate variability across runs. All control commands and errors converge toward zero, indicating robust trajectory tracking and control convergence from various initial conditions. Note the striking similarity in drone control and distance error to target between the complex analytical teacher and the small ConvNet student.}
    \label{fig:real-evolution-comparison}
\end{figure*}

\subsection{Mission Termination Conditions}
\label{subsed:experiments-termination-conditions}
We implemented three termination conditions, all using frames of size $640 \times 360$ pixels. First, a hard timeout of 75 seconds triggered automatic landing if the drone failed to correct its pose within this time frame. Second, a soft condition was activated when the median of the last 5 errors was $\leq 80$ pixels and all velocity commands remained at 0 for 3 consecutive seconds, indicating acceptable error levels but potential entrapment in a local minimum. Third, a hard condition is triggered when the median of the last 5 errors was $\leq 40$ pixels for 3 consecutive seconds, signifying highly acceptable error levels where NSER IBVS generated minimal commands, producing negligible movement. The error is computed as the L2 norm for the vector that contains the raw pixel values.

To independently evaluate the student model, without requiring IBVS and the two neural networks (YOLOv11 and the Mask Splitter), we empirically derived the student termination condition by analyzing the statistics from teacher method evaluations. We examined the velocity command values within the last three seconds of each evaluation run. This analysis indicated that velocity commands were typically within the absolute range of 0 to 1 for most of our tests. Therefore, we established this as our termination condition for the simulated student: if the absolute value of the last 5 commands was $\leq 1$ for 3 consecutive seconds, the drone would land and terminate the testing process.

For the real-world environment, testing followed a similar method, but with fewer data points. We initially tested 10 flight sequences for each pose as in the simulator (Figure \ref{fig:drone-starting-poses}), yielding 80 real-world evaluations having $43963$ frames. The student model for real-world deployment was pre-trained on the 240 simulated runs and subsequently fine-tuned on the real-world scenes. The termination condition for drone landing in the real world was identical to that used in the simulator.

\subsection{Evaluation Metrics}
\label{subsec:experiments-metrics}
We compared the performance of teacher and student models using several metrics. The primary metrics were \textbf{error norm} and \textbf{Intersection over Union} (IoU) between the goal points and current points in the image during the last three seconds before landing. Furthermore, we measured the total \textbf{flight duration} and \textbf{distance} traveled from the point where tracking began (drone take-off with camera angled 45 degrees downward) until the point of landing. For the student method, these metrics were computed offline.

\subsection{Results and Discussion}
\label{sec:discussions-results}
The results demonstrate that the student model achieves significantly improved tracking accuracy compared to the teacher method. In simulation, the student model achieves a mean error norm of $14.261$ pixels compared to the teacher $29.756$ pixels, representing a $52\%$ improvement in tracking precision. The IoU metric shows similar improvements, with the student achieving $0.752$ compared to the teacher $0.522$, indicating better object coverage and localization \ref{tab:teacher-student-metricwise}.

\begin{table*}[htpb]
    \centering
    \small
    \setlength{\tabcolsep}{4pt}
    \renewcommand{\arraystretch}{0.9}
    \begin{tabular}{c c c c c c c c}
         &  & \multicolumn{1}{c}{\makecell{Flight SIM \\ (distance(m) / time(s)) $\downarrow$}} & \multicolumn{1}{c}{\makecell{Norm \\ Error SIM (px) $\downarrow$}} & \multicolumn{1}{c}{IoU SIM $\uparrow$} & \multicolumn{1}{c}{\makecell{Flight \\ (distance(m) / time(s)) $\downarrow$}} & \multicolumn{1}{c}{\makecell{Norm \\ Error (px) $\downarrow$}} & \multicolumn{1}{c}{IoU $\uparrow$} \\
        \addlinespace
        \cmidrule(lr){3-5} \cmidrule(lr){6-8} \multirow{2}{*}{Up-Left} & Teacher & 5.312 / \textbf{23.466} & 29.256 & 0.530 & 5.798 / \textbf{36.721} & 30.134 & 0.620 \\
                               & Student & \textbf{5.193} / 24.164 & \textbf{13.319} & \textbf{0.759} & \textbf{5.735} / 43.334 & \textbf{28.600} & \textbf{0.6263} \\
        \addlinespace
        \multirow{2}{*}{Up-Right} & Teacher & \textbf{5.675 / 24.226} & 31.800 & 0.503 & \textbf{5.622 / 41.581} & 31.499 & 0.621\\
                               & Student & 6.064 / 28.298 & \textbf{13.172} & \textbf{0.766} & 5.716 / 45.885 & \textbf{22.802} & \textbf{0.6919}\\
        \addlinespace
        \multirow{2}{*}{Front-Left} & Teacher & 6.196 / \textbf{27.315} & 30.706 & 0.517 & 6.493 / \textbf{37.535} & \textbf{28.54} & \textbf{0.611}\\
                               & Student & \textbf{6.041} / 27.917 & \textbf{13.430} & \textbf{0.758} & \textbf{6.490} / 47.238 & 33.981 & 0.560\\
        \addlinespace
        \multirow{2}{*}{Front-Right} & Teacher & \textbf{6.846 / 32.358} & 32.608 & 0.488 & \textbf{6.197 / 37.166} & 33.15 & \textbf{0.658}\\
                               & Student & 7.043 / 35.535 & \textbf{18.028} & \textbf{0.718} & 6.316 / 46.363 & \textbf{31.035} & 0.627\\
        \addlinespace
        \multirow{2}{*}{Left} & Teacher & 4.177 / \textbf{20.228} & 31.453 & 0.519 & \textbf{4.559 / 29.921} & \textbf{28.015} & 0.629\\
                               & Student & \textbf{4.089} / 20.481 & \textbf{13.243} & \textbf{0.762} & 5.065 / 43.841 & 30.853 & \textbf{0.6482} \\
        \addlinespace
        \multirow{2}{*}{Right} & Teacher & \textbf{4.317 / 19.637} & 31.137 & 0.494 & 4.831 / \textbf{41.409} & \textbf{32.423} & \textbf{0.612}\\
                               & Student & 4.518 / 21.987 & \textbf{13.798} & \textbf{0.759} & \textbf{4.811} / 57.245 & 43.672 & 0.5 \\
        \addlinespace
        \multirow{2}{*}{Down-Left} & Teacher & 2.779 / 15.988 & 28.473 & 0.518 & 4.384 / \textbf{31.622} & \textbf{28.00} & \textbf{0.611}\\
                               & Student & \textbf{2.777 / 14.900} & \textbf{13.257} & \textbf{0.763} & \textbf{4.326} / 41.044 & 39.531 & 0.5253 \\
        \addlinespace
        \multirow{2}{*}{Down-Right} & Teacher & \textbf{2.893 / 13.667} & 22.618 & 0.606 & 4.137 / \textbf{33.523} & \textbf{27.89} & \textbf{0.654}\\
                               & Student & 3.145 / 17.035 & \textbf{15.839} & \textbf{0.728} & \textbf{3.938} / 38.001 & 36.195 & 0.5478 \\
        \bottomrule
        \addlinespace
        \multirow{2}{*}{Mean} & Teacher & \textbf{4.774 / 22.111} & 29.756 & 0.522 & \textbf{5.253 / 36.185} & \textbf{29.956} & \textbf{0.627}\\
                               & Student & 4.859 / 23.790 & \textbf{14.261} & \textbf{0.752} & 5.300 / 45.369 & 33.334 & 0.591 \\
        \bottomrule
    \end{tabular}
    \caption{Teacher-student comparison across different starting positions. The left side shows results in the simulator; the right side shows results on flights in the real world. Metrics include total flight distance/time, final norm error in pixels (L2 norm of the error vector for all 4 corner points combined), and final IoU (L2 norm and IOU are computed in the last 3 secs of the flight). The teacher is slightly faster in flight time. However, the student is $11\times$ faster in computation time (see Tab. \ref{tab:nn-timing-eval}). The student is slightly more accurate (lower errors at destination than the teacher) in simulation, where it was trained more, but it is slightly less accurate in real-world, where it was trained on far fewer flights.}
    \label{tab:teacher-student-metricwise}
\end{table*}

\begin{figure}
    \centering
    \includegraphics[width=\linewidth]{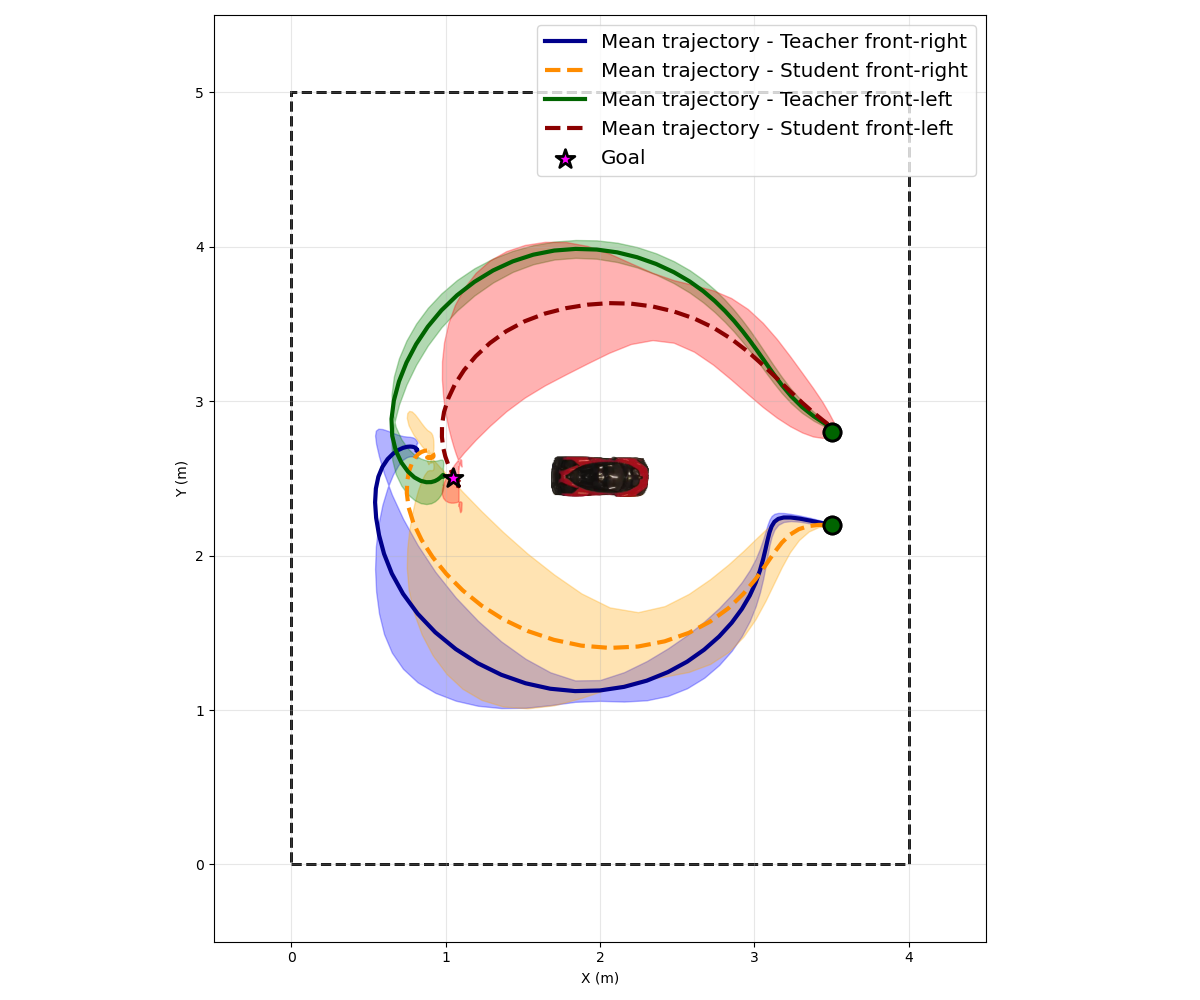}
    \caption{Trajectories of teacher and student simulation flights, with mean and standard deviation for 2 starting poses (green circles: front-left and front-right). The solid blue and green lines represent mean paths of teacher NSER IBVS method, while dashed orange and red lines show student paths. Shaded regions indicate trajectory variability across runs. The star represents the goal pose. Note that the student displays more path variation, but it has a shorter average path than the teacher.}
    \label{fig:trajectory-ibvs}
\end{figure}

The trajectory evolution analysis reveals that both methods successfully converge toward the target, but with different characteristics. The teacher method shows more conservative command profiles with gradual convergence, while the student method demonstrates more direct approaches with faster error reduction rates shown in Figure \ref{fig:trajectory-ibvs}.

To compare the computational efficiency of our proposed student network with and without segmentation against the baseline NSER IBVS, we conducted a runtime benchmarking experiment on a sequence of $640 \times 360$ resolution real-world frames. Each evaluator processed the same set of input frames independently, and we measured per-frame inference time over 30 repeated trials to account for variability due to system load and memory effects. For each trial, we performed warm-up iterations to avoid initialization bias and explicitly released memory between runs. We reported inference times in milliseconds and the FPS rate. The results are summarized in the Table \ref{tab:nn-timing-eval}. 

\begin{table}[t]
\centering
\footnotesize
\caption{Computation times (in milliseconds)
over 30 trials. The small 1.7M params student ConvNet is $11\times$ faster than the teacher.}
\label{tab:nn-timing-eval}
\begin{tabular}{lrrrrrr}
\toprule
\textbf{Evaluator} & \textbf{Avg} & \textbf{Std} & \textbf{Med} & \textbf{Min} & \textbf{Max} & \textbf{FPS} \\
\midrule
NSER IBVS & 20.69 & 7.63 & 24.56 & 6.45 & 82.55 & 48.30 \\
Student  & \textbf{1.85} & 0.93 & 1.84 & 1.79 & 235.64 & \textbf{540.8} \\
\bottomrule
\end{tabular}
\end{table}

The results show effective sim-to-real transfer, though with some performance degradation as expected. In real-world scenarios, the student model maintains competitive tracking accuracy while the teacher method shows more consistent performance across different environments. The fine-tuning approach using 80 real-world scenes proves effective for domain adaptation, enabling successful deployment in practical scenarios.

\section{Final Remarks and Conclusions}
\label{sec:conclusion}
This work introduces the first self-supervised neuro-analytic system for visual servoing of drones, through efficient knowledge distillation: a small ConvNet student learns from the proposed, numerically stable analytical teacher (NSER-IBVS) to fly the drone from different start poses to an end pose, which fully cover the relative pose difference on the trigonometric circle. The system is fully unsupervised, eliminating any human intervention, supervised pretraining or external sensors, unlike any other method in the literature. The student net, which is only 1.7M parameters and easily deployable on embedded devices, is $11\times$ computationally faster ($529.77$ FPS) than the teacher pathway ($48.3$ FPS), while remaining equally accurate and fast in flight, in both simulation and real world tests. Learning on real-world cases is limited to only a few flights ($80$), to further reduce the costs, through an efficient transfer between pretraining in the digital twin simulation environment and the real-world scene.

Our novel two-stage segmentation pipeline, combining YOLOv11 with a specialized U-Net-based mask splitter, provides robust anterior-posterior vehicle segmentation without requiring explicit geometric models or fiducial markers. The self-supervised nature of this approach, validated through simulation-to-reality transfer learning and testing on a Parrot Anafi drone platform, demonstrates the practical feasibility of fully autonomous indoor operations with minimal hardware infrastructure requirements. 

Experimental validation across distinct drone poses fully covering the trigonometric circle confirms the effectiveness of our unsupervised approach. While the student network exhibits slightly higher pixel errors in the real world (due to very limited real training flights) when compared to the teacher (mean error of $29.956$ vs $33.334$ pixels), it maintains acceptable IoU performance ($0.627$ vs $0.591$) and achieves faster convergence times. Notably, in the simulated environment where more extensive self-supervised training occurs, the system demonstrates superior convergence times with reduced errors, validating the cost-effectiveness of simulation-based training compared to expensive real-world data collection. Computational efficiency gains make our approach particularly valuable for applications requiring real-time performance on resource-constrained platforms, establishing a new paradigm for autonomous visual servoing without human supervision or marker dependencies.

\section*{Acknowledgments} 
This work is supported in part by projects "Romanian Hub for Artificial Intelligence - HRIA", Smart Growth, Digitization and Financial Instruments Program, 2021-2027 (MySMIS no. 334906), European Health and Digital Executive Agency (HADEA), under the powers delegated by the European Commission, through the DIGITWIN4CIUE project with grant agreement No. 101084054., and "European Lighthouse of AI for Sustainability - ELIAS", Horizon Europe program (Grant No. 101120237).
{
\small
\bibliographystyle{ieeenat_fullname}
\bibliography{main}
}

\end{document}